\documentclass[10pt,twocolumn]{article}

\usepackage[letterpaper,margin=0.85in,columnsep=0.3in]{geometry}
\usepackage{times}
\usepackage[utf8]{inputenc}
\usepackage[T1]{fontenc}
\usepackage{amsmath,amssymb}
\usepackage{graphicx}
\usepackage{booktabs}
\usepackage{multirow}
\usepackage{algorithm}
\usepackage{algpseudocode}
\usepackage{float}
\usepackage{microtype}
\usepackage{url}
\usepackage[round]{natbib}
\usepackage{balance}   
\usepackage{adjustbox} 
\usepackage{flafter}
\usepackage{placeins}   
\usepackage{caption}
\usepackage{tikz}
\usetikzlibrary{positioning,arrows.meta,fit,backgrounds}
\usepackage[colorlinks=true,linkcolor=black,citecolor=black,urlcolor=black]{hyperref}

\tikzset{
  zbox/.style   = {draw, rounded corners=2pt, align=center, font=\small,
                   inner sep=4pt, minimum height=7mm},
  zin/.style    = {draw, rounded corners=2pt, align=center, font=\scriptsize,
                   fill=black!5, inner sep=3pt},
  zarr/.style   = {-{Stealth[length=5pt]}, thick},
  zloop/.style  = {draw, dashed, rounded corners=4pt, inner sep=7pt},
}

\newcommand{\zaps}{\textsc{ZAPS}}
\newcommand{\pfit}{\textsc{ProxyFit}}
\newcommand{\patk}{\mbox{P@100}}
\newcommand{\pmstd}[1]{{\scriptstyle\,\pm\,#1}}

\title{\vspace{-1.2em}ZAPS: Zero-Cost Active Proxy Search\\for Neural
Architecture Search\vspace{-0.3em}}

\author{
Hassan Touayouch \\ ENSICAEN
\and
Rabie Najem \\ ISIA Lab, Universit\'e de Mons
\and
Mohammed Benjelloun \\ ISIA Lab, Universit\'e de Mons
}
\date{}

\begin{document}

\makeatletter
\twocolumn[
  \begin{@twocolumnfalse}
  \maketitle
  \begin{quote}
  \begin{center}\textbf{Abstract}\end{center}
  \vspace{-0.3em}
Neural Architecture Search (NAS) automates network design, but evaluating a single
candidate requires training it to convergence, making exhaustive search intractable.
Zero-cost proxies estimate architecture quality at initialization in seconds, yet
a single proxy is noisy, and combining several does not straightforwardly help:
proxies are strongly correlated, so naive aggregation compounds their shared
errors instead of averaging them out. Existing methods exploit either proxy
signals or architectural topology---never both within a single active-learning
framework.
We introduce \zaps{} (\emph{Zero-cost Active Proxy Search}), a four-stage pipeline
that closes this gap. \zaps{} (i)~selects a compact, non-redundant proxy subset
offline via \pfit{}, a greedy anti-redundancy criterion; (ii)~seeds the search with
a hybrid K-means strategy that balances exploitation and exploration;
(iii)~re-selects proxies at every iteration by a bootstrapped vote as the
labeled set grows; and (iv)~ranks candidates with an XGBoost ensemble trained
jointly on proxy ranks and one-hot topological encodings, queried through an
Upper Confidence Bound (UCB) acquisition function.
On NAS-Bench-201 under a budget of $B=200$ evaluations, \zaps{} recovers
\textbf{52.3\,\%} of the true top-100 architectures on CIFAR-10 and
\textbf{65.8\,\%} on CIFAR-100, ahead of every baseline we consider---Random
Search, Local Search, REA, BANANAS and TPE---and, on CIFAR-10, with less than
half the run-to-run standard deviation of the strongest of them. The advantage
is largest where evaluations are scarce: on NAS-Bench-201 it narrows as the
budget grows, whereas on the harder NAS-Bench-101, which no method comes close
to saturating, it widens instead. All methods are scored by a single criterion:
how much of the true top-100 lies among the architectures they actually
evaluated.
  \end{quote}
  \vspace{1.2em}
  \end{@twocolumnfalse}
]
\makeatother

\section{Introduction}

Neural Architecture Search (NAS) automates the design of high-performing networks
by searching a predefined space of candidate architectures~\citep{elsken2019survey}.
Its dominant cost is evaluation: obtaining the accuracy of a single candidate
requires training it to convergence, so covering a realistic search space demands
thousands of GPU hours. Reducing the number of such evaluations---or replacing them
with cheaper signals---is therefore the central problem in practical NAS.

\paragraph{Zero-cost proxies and their limits.}
Zero-cost proxies~\citep{mellor2021naswot,tanaka2020synflow} score an architecture
at initialization from a single forward or backward pass, cutting evaluation time
from hours to seconds. Each proxy probes a distinct network property---expressivity,
gradient flow, parameter count---which suggests combining several of them to obtain
a broader signal. In practice, however, proxies are far from independent:
on NAS-Bench-201 / CIFAR-10, \texttt{flops} and \texttt{params} correlate at
$\rho=0.996$. Combining proxies naively therefore duplicates information and
amplifies correlated noise rather than enriching the signal.

\paragraph{The gap.}
Existing methods fall into two disjoint families.
\emph{Proxy-based} methods rely on a single score~\citep{mellor2021naswot,
tanaka2020synflow,li2023zico} or aggregate several through a
uniform vote~\citep{rankproduct}; they neither filter redundancy nor exploit the
topological structure of the architecture.
\emph{Surrogate-based} methods~\citep{white2021bananas,nasbowl} learn a predictor
over architecture encodings but ignore zero-cost signals entirely, and consequently
need substantially larger labeled sets before their surrogate becomes accurate.
To our knowledge, no prior work simultaneously (a)~selects a non-redundant,
high-relevance proxy subset and (b)~fuses it with architectural topology inside an
active-learning loop.

\paragraph{Our approach.}
We propose \zaps{}, which is built on the hypothesis that proxy scores and
architectural structure are \emph{complementary} rather than interchangeable:
proxies measure functional quality at initialization, whereas a one-hot topology
encoding describes the computation graph that determines expressive capacity.
\zaps{} feeds both into a single surrogate and queries it through a UCB acquisition
function, so that every evaluation spent is chosen to be maximally informative.
Our ablation (Section~\ref{sec:ablation}) confirms that each pipeline component is
necessary: removing the anti-redundancy filter costs $11$--$13$\,pp of \patk{}, and
removing the structured initialization costs $16$--$23$\,pp while multiplying
the run-to-run standard deviation by two to three.

\paragraph{Contributions.}
\begin{enumerate}
  \item \textbf{\pfit{}} (Section~\ref{sec:proxyfit}): an offline greedy algorithm
        that extracts a compact proxy subset maximizing correlation with
        ground-truth accuracy under an explicit anti-redundancy constraint.
  \item \textbf{Hybrid K-means initialization} (Section~\ref{sec:kmeans}): a
        two-phase seeding strategy that balances exploitation of proxy-favored
        regions against coverage of the remaining search space.
  \item \textbf{Bootstrapped online re-selection} (Section~\ref{sec:mrmr}): a
        dynamic rule that adapts the active proxy subset to the growing labeled
        set, stabilized by a bootstrap majority vote and halted by an MRMR
        stopping criterion.
  \item \textbf{A proxy-augmented surrogate} (Section~\ref{sec:xgboost}): an
        XGBoost ensemble trained on concatenated proxy ranks and one-hot topology,
        driving a UCB acquisition function.
  \item \textbf{An empirical study} (Section~\ref{sec:experiments}) over $200$
        independent seeds on two search spaces, NAS-Bench-201 and NAS-Bench-101,
        establishing that \zaps{} leads every baseline we consider at moderate
        budget while being markedly more stable across seeds. Every baseline is
        audited against its source publication and a behavioral check before use.
\end{enumerate}

\FloatBarrier
\section{Related Work}

\paragraph{Zero-cost proxies.}
NASWOT~\citep{mellor2021naswot} scores architectures by the log-determinant of a
kernel matrix built from the binary activation patterns of their ReLU units, which
measures how distinctly an untrained network separates inputs.
Synflow~\citep{tanaka2020synflow} propagates a data-independent gradient signal that
avoids layer collapse. SNIP~\citep{lee2019snip} measures connection saliency, and
ZiCo~\citep{li2023zico} quantifies gradient stability across layers.
Despite their efficiency, each of these scores captures a single facet of
architectural quality and its predictive power varies considerably across search
spaces and datasets. NAS-Bench-Suite-Zero~\citep{nasbenchsuitezero} standardizes this
landscape by releasing precomputed values for $13$ proxies over several benchmarks;
we build directly on this resource.

\paragraph{Proxy combination.}
\citet{rankproduct} ensemble zero-cost proxies by a majority vote among three
fixed proxies, weighting each equally and without discounting the redundancy
between them.
EZNAS~\citep{eznas} evolves symbolic composite operators over proxy statistics, and
LPZero~\citep{lpzero} transfers this idea to language-model search spaces.
UP-NAS~\citep{superproxy} learns a multi-proxy estimator that consolidates several
zero-cost proxies into a unified score, but the estimator itself must be trained on
labeled architectures.
All of these produce a \emph{single scalar score}; none feeds the selected proxies
into a learned surrogate alongside structural features.
\pfit{} differs on both counts: it selects proxies for joint relevance and mutual
complementarity, and passes the resulting subset to a surrogate that also consumes
architectural topology.

\paragraph{Surrogate-based NAS.}
BANANAS~\citep{white2021bananas} represents architectures by path encodings and fits
an ensemble surrogate for Bayesian optimization; NASBOWL~\citep{nasbowl} employs a
graph-kernel Gaussian process at higher computational cost. Both learn exclusively
from architecture encodings and require comparatively large labeled sets before the
surrogate becomes reliable---a regime our budgets deliberately exclude.
\zaps{} augments this family by injecting zero-cost proxy ranks into the surrogate's
feature space; to our knowledge it is the first method to combine the two signal
types in a single surrogate under a strict evaluation budget.

\paragraph{Active learning and classical NAS.}
Active-learning approaches to NAS~\citep{activenas} choose which architectures to
label next based on surrogate uncertainty, again using encodings alone.
Classical NAS relies on reinforcement learning~\citep{zoph2017neural} or regularized
evolution~\citep{real2019regularized}, at a cost of thousands of GPU days;
DARTS~\citep{liu2019darts} reduces this through a differentiable relaxation but does
not apply to the tabular benchmarks studied here.
\zaps{} combines UCB-driven active learning with a proxy-augmented surrogate,
inheriting the sample efficiency of the former and the cheap prior of the latter.

\FloatBarrier
\section{Method}
\label{sec:method}

\paragraph{Problem setting.}
Let $\mathcal{A}$ be a discrete architecture search space and let
$f:\mathcal{A}\to\mathbb{R}$ denote true validation accuracy, accessible only
through full training. Given a budget $B$ of such evaluations, a method produces a
set $\mathcal{E}_B\subseteq\mathcal{A}$ of $B$ architectures it chose to evaluate.
Our primary objective is to recover $\mathcal{T}_{100}$, the $100$ best
architectures in $\mathcal{A}$, and we measure how much of it a method actually
reached:
\begin{equation}
  \patk{} \;=\; \tfrac{1}{100}\,\bigl|\,\mathcal{E}_B\cap\mathcal{T}_{100}\,\bigr|
  \;\times\;100\,\%.
  \label{eq:p100}
\end{equation}
Scoring on $\mathcal{E}_B$ rather than on a predicted ranking is what makes the
comparison meaningful: search methods such as Random Search or REA produce no
ranking over $\mathcal{A}$, so a predicted top-100 is undefined for them, and
crediting surrogate methods for architectures they never trained would compare
unlike quantities. Every method here is scored by Eq.~\ref{eq:p100}.
As a secondary criterion we report \emph{regret},
$R = (f^{\star}-f_{\mathrm{best}})/f^{\star}\times 100\,\%$, where
$f^{\star}=\max_{a\in\mathcal{A}} f(a)$ and
$f_{\mathrm{best}}=\max_{a\in\mathcal{E}_B} f(a)$.
\patk{} thus measures how much of the top region a method reaches; regret measures
how good its single best find is.

\begin{figure}[H]
\centering
\includegraphics[width=\linewidth]{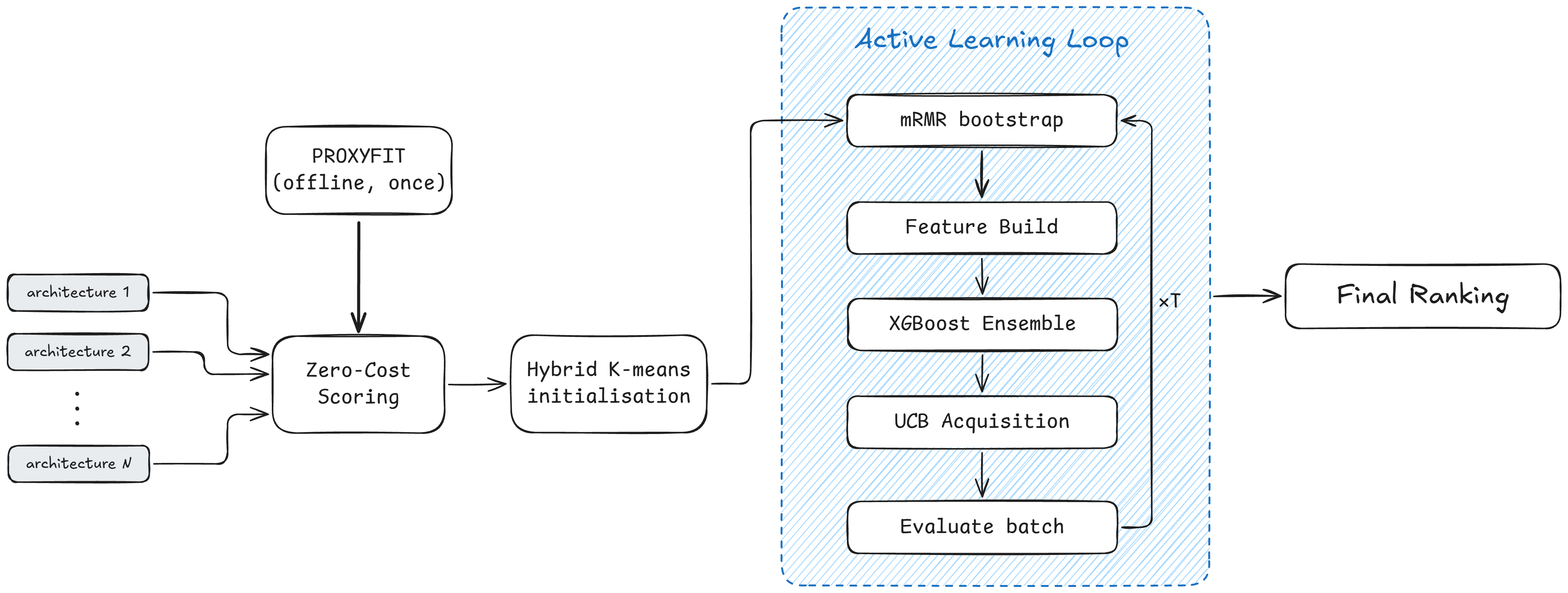}
\caption{The four stages of \zaps{}. \pfit{} runs once offline and returns a proxy
subset; the hybrid K-means seeds the labeled set; re-selection, surrogate fitting
and UCB querying then alternate for $T$ iterations, each one spending part of the
budget on newly evaluated architectures.}
\label{fig:pipeline}
\end{figure}

\paragraph{Pipeline overview.}
\zaps{} comprises four stages (Figure~\ref{fig:pipeline}):
\textbf{(1)}~offline proxy selection with \pfit{};
\textbf{(2)}~hybrid K-means initialization of the labeled set;
\textbf{(3)}~dynamic proxy re-selection by a bootstrapped vote; and
\textbf{(4)}~UCB-guided querying of an XGBoost ensemble.
Stage~1 runs once; stages~3 and~4 alternate inside the active-learning loop
(Algorithm~\ref{alg:zaps}).

\subsection{ProxyFit: Anti-Redundant Proxy Selection}
\label{sec:proxyfit}

Let $P$ denote the pool of $|P|=13$ available zero-cost proxies
(Table~\ref{tab:proxies}). Supplying all of them to the surrogate injects correlated
noise and dilutes the informative directions of the feature space.
\pfit{} instead extracts a compact subset $\Pi\subseteq P$ of size $K$: it traverses
the proxies in decreasing order of $|\rho(p,f)|$---their Spearman correlation with
accuracy---and admits a candidate only when its correlation with every
already-selected proxy stays below a threshold $\tau$
(Algorithm~\ref{alg:proxyfit}). Relevance thus determines the traversal order,
while redundancy acts as an admission filter.

\begin{figure}[H]
\centering
\includegraphics[width=\linewidth]{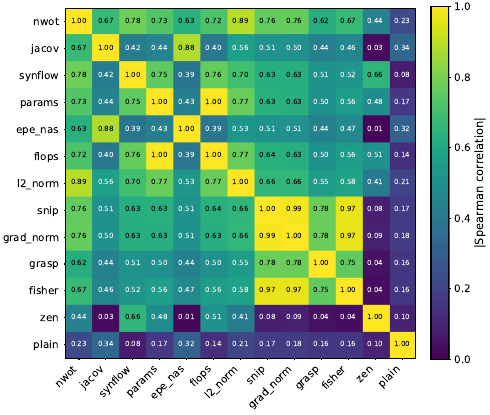}
\caption{Absolute Spearman correlation between proxy pairs on NAS-Bench-201 /
CIFAR-10, with proxies ordered by decreasing relevance---the order in which \pfit{}
examines them. Redundancy is assessed on $|\rho|$, since two strongly
anti-correlated proxies are as redundant as two strongly correlated ones. Highly
correlated pairs such as \texttt{flops}--\texttt{params} ($|\rho|=0.996$) are the
redundancies \pfit{} is designed to eliminate.}
\label{fig:proxy_corr}
\end{figure}

On CIFAR-10 with $\tau=0.85$ and $K=6$, \pfit{} returns
$\{\texttt{nwot},\allowbreak\ \texttt{jacov},\allowbreak\ \texttt{synflow},\allowbreak\
\texttt{params},\allowbreak\ \texttt{snip},\allowbreak\ \texttt{grasp}\}$.
Notably it rejects \texttt{flops}, whose correlation with the already-admitted
\texttt{params} reaches $0.996$; Figure~\ref{fig:proxy_corr} shows the full
correlation structure that drives these decisions.
\paragraph{On the use of ground-truth accuracies.}
\pfit{} computes $\rho(p,f)$ over the whole space, so as stated it consumes labels
that a search method operating under budget $B$ would not have. We make this
explicit rather than leave it implicit, and we neutralize it in two ways.
First, \pfit{} is an \emph{offline, one-off} calibration whose output is a list of
six proxy names---not a ranking of architectures---so it is reusable across every
subsequent search on that space. Second, and decisively, the calibration turns out
not to matter: substituting a subset calibrated on a \emph{different} dataset
leaves performance unchanged within $5.2$\,pp of \patk{} at $B\ge 200$
(Section~\ref{sec:transferability}). The reported gains therefore do not stem from
privileged access to target labels, and the headline conclusions of
Section~\ref{sec:main} hold with a transferred subset.

\begin{algorithm}[tb]
\caption{\pfit{}: greedy anti-redundant proxy selection}
\label{alg:proxyfit}
\begin{algorithmic}[1]
\Require Proxy pool $P$, redundancy threshold $\tau=0.85$, target size $K=6$
\Ensure Selected subset $\Pi\subseteq P$
\State Sort $P$ by $|\rho(p,f)|$ in decreasing order
\State $\Pi \leftarrow \emptyset$
\For{each $p$ in sorted order}
  \If{$\max_{q\in\Pi}|\rho(p,q)| < \tau$}
    \State $\Pi \leftarrow \Pi\cup\{p\}$
  \EndIf
  \If{$|\Pi| = K$} \textbf{break} \EndIf
\EndFor
\State \Return $\Pi$
\end{algorithmic}
\end{algorithm}

\begin{table}[H]
\centering
\footnotesize
\setlength{\tabcolsep}{5pt}
\caption{The $13$ zero-cost proxies of NAS-Bench-Suite-Zero, grouped by family.
The last column gives the range of Spearman $\rho$ with true accuracy on
NAS-Bench-201 / CIFAR-10.}
\label{tab:proxies}
\adjustbox{max width=\linewidth}{%
\begin{tabular}{llc}
\toprule
\textbf{Family} & \textbf{Proxies} & \textbf{Spearman} $\rho$ \\
\midrule
Deterministic & \texttt{params}, \texttt{flops}, \texttt{l2\_norm}, \texttt{synflow} & $0.68$--$0.73$ \\
Activation    & \texttt{nwot}, \texttt{jacov}, \texttt{epe\_nas}, \texttt{zen}, \texttt{plain} & $-0.27$--$0.78$ \\
Gradient      & \texttt{fisher}, \texttt{snip}, \texttt{grad\_norm}, \texttt{grasp} & $0.51$--$0.60$ \\
\bottomrule
\end{tabular}}
\end{table}

\subsection{Hybrid K-means Initialization}
\label{sec:kmeans}

Surrogate quality under a tight budget depends critically on the initial labeled set
$\mathcal{L}_0$. Uniform random sampling wastes evaluations on uninformative regions,
whereas selecting purely by proxy score confines the surrogate to a narrow,
proxy-biased slice of the space and leaves it unable to correct proxy errors.
We therefore allocate the initialization budget $n_0$ across two complementary
phases:
\begin{itemize}
  \item \textbf{Exploitation phase} ($n_0/2$ architectures): K-means restricted to
        the top $30\,\%$ of $\mathcal{A}$ ranked by aggregate proxy score, retaining
        the architecture closest to each centroid. This concentrates labels where
        high-accuracy architectures are likely to lie, while clustering prevents the
        sample from collapsing onto a single mode.
  \item \textbf{Coverage phase} ($n_0/2$ architectures): K-means over the entire
        space minus the points already chosen, again retaining one representative
        per cluster. Because this pass is unrestricted, it reaches the regions the
        proxies rank poorly---precisely where the surrogate must learn to correct
        them---while its own clustering keeps the sample spread out.
\end{itemize}
The two phases are thus not disjoint: the second clusters over the whole space
minus the points already chosen, so it ranges over the proxy-favored region as
well rather than being confined to its complement. The exploitation phase acts as
a deliberate over-sampling of that region rather than as a partition of the space.
Both phases cluster in proxy-rank space. We set $n_0=\max(10,\lfloor B/4\rfloor)$
($50$ architectures at $B=200$), leaving the remainder for the active loop; when
$B-n_0$ is not divisible by $T$, the floor leaves at most $T-1$ evaluations of the
budget unspent.

\subsection{Bootstrapped Online Re-Selection}
\label{sec:mrmr}

The subset chosen offline by \pfit{} is calibrated on the search space as a whole,
but the relevance and redundancy of a proxy \emph{conditioned on the labeled set}
shift as $\mathcal{L}_t$ grows. We therefore re-select an active subset $\Pi_t$ at
every iteration, applying the same relevance-ordered, redundancy-filtered traversal
as \pfit{} but with all correlations re-estimated on $\mathcal{L}_t$ alone.

Two elements differ from the offline stage. First, we add a stopping rule based on
the Maximum Relevance Minimum Redundancy score of a candidate $p$ against the
partial selection $S$ built so far,
\begin{equation}
  J(p \mid S) \;=\; |\rho(p,f)| \;-\; \frac{1}{|S|}\sum_{q\in S}|\rho(p,q)|,
  \label{eq:mrmr}
\end{equation}
which halts the traversal once the leading candidate contributes more redundancy
than relevance ($J(p \mid S)<0$), provided $|S|\ge k_{\min}$.
Second, because correlations estimated on a small $\mathcal{L}_t$ are unstable, we
repeat the traversal on $B_{\mathrm{boot}}=3$ bootstrap resamples and keep the
proxies with the most votes (Algorithm~\ref{alg:mrmr}). The target size $k$ grows
with the labeled set up to a ceiling, so that model capacity tracks the available
evidence.

\begin{algorithm}[tb]
\caption{Bootstrapped online re-selection}
\label{alg:mrmr}
\begin{algorithmic}[1]
\Require Labeled set $\mathcal{L}_t$, proxy pool $P$, $B_{\mathrm{boot}}=3$,
         $k_{\min}=4$, $k_{\max}=10$, threshold $\tau=0.85$
\Ensure Active subset $\Pi_t\subseteq P$
\State $k \leftarrow
\max\bigl(k_{\min},\,\min(k_{\max},\,\lfloor|\mathcal{L}_t|/15\rfloor+4)\bigr)$
\State $\mathrm{votes}[p]\leftarrow 0$ for all $p\in P$
\For{$b=1$ \textbf{to} $B_{\mathrm{boot}}$}
  \State $\mathcal{L}^{b}\leftarrow$ bootstrap resample of $\mathcal{L}_t$;
         estimate all $\rho$ on $\mathcal{L}^{b}$
  \State Sort $P$ by $|\rho(p,f)|$ decreasing; $\Pi^{b}\leftarrow\{$first proxy$\}$
  \For{each remaining $p$ in sorted order}
    \State \textbf{stop} if $|\Pi^{b}| = k_{\max}$,
           or if $J(p \mid \Pi^{b})<0$ and $|\Pi^{b}|\ge k_{\min}$
    \State $\Pi^{b}\leftarrow\Pi^{b}\cup\{p\}$
           \textbf{ if } $\max_{q\in\Pi^{b}}|\rho(p,q)| < \tau$
  \EndFor
  \State $\mathrm{votes}[p]\mathrel{+}=1$ for each $p\in\Pi^{b}$
\EndFor
\State \Return the $k$ proxies with the highest vote counts
\end{algorithmic}
\end{algorithm}

\subsection{Proxy-Augmented Surrogate and UCB Acquisition}
\label{sec:xgboost}

\paragraph{Feature representation.}
Each architecture $a$ is represented by
$\phi(a)=\bigl[\hat{\mathbf{r}}(a);\,\mathbf{e}(a)\bigr]$, the concatenation of
\emph{(i)}~normalized proxy ranks $\hat{\mathbf{r}}(a)\in[0,1]^{|\Pi_t|}$, computed
over the whole space so that a single outlying proxy value cannot dominate the
feature, and \emph{(ii)}~a binary encoding $\mathbf{e}(a)$ of the cell's structure:
on NAS-Bench-201, one-hot over the operation assigned to each of the six cell
edges ($6\times5=30$ dimensions); on NAS-Bench-101, whose operations sit on nodes
rather than edges, the strict upper triangle of the $7\times7$ adjacency matrix
($21$ bits) concatenated with a one-hot operation code per node
($7\times5=35$), for $56$ in total. The first block carries a cheap functional
prior available before any training; the second carries the exact discrete structure
the proxies can only summarize. Their concatenation is what distinguishes \zaps{}
from both proxy-only and encoding-only methods; to our knowledge no prior zero-cost
method supplies both signal types to a single regressor.

\paragraph{Ensemble and acquisition.}
We fit three XGBoost~\citep{chen2016xgboost} regressors on independent bootstrap
resamples of $\mathcal{L}_t$, yielding for every unevaluated architecture a
predictive mean $\mu(a)$ and a dispersion $\sigma(a)$ that serves as an uncertainty
estimate. The next batch of $b=\lfloor(B-n_0)/T\rfloor$ architectures maximizes the
Upper Confidence Bound
\begin{equation}
  \alpha(a) \;=\; \mu(a) \;+\; \beta\,\sigma(a),
  \qquad \beta = 0.5,
  \label{eq:ucb}
\end{equation}
balancing exploitation of high predicted accuracy against exploration of regions
where the ensemble disagrees. The loop repeats for $T$ iterations until the budget
is exhausted, re-selecting proxies and refitting the ensemble each time;
Algorithm~\ref{alg:zaps} states the assembled procedure.

\begin{algorithm}[H]
\caption{\zaps{}: full active-learning loop}
\label{alg:zaps}
\begin{algorithmic}[1]
\Require Search space $\mathcal{A}$, budget $B$, iterations $T=2$,
         exploration weight $\beta$
\Ensure Predicted top-100 architectures
\State $\Pi_{\mathrm{init}} \leftarrow \pfit{}(P)$ \Comment{offline, once}
\State $n_0 \leftarrow \max(10,\lfloor B/4\rfloor)$; \;
       $b \leftarrow \max\bigl(1,\lfloor (B-n_0)/T\rfloor\bigr)$
\State $\mathcal{L}_0 \leftarrow$ hybrid K-means initialization ($n_0$
       architectures) using $\Pi_{\mathrm{init}}$; evaluate $f$ on $\mathcal{L}_0$
\For{$t=0$ \textbf{to} $T-1$}
  \State $\Pi_t \leftarrow$ bootstrapped re-selection on $\mathcal{L}_t$
         (Algorithm~\ref{alg:mrmr})
  \State Build $\phi(a)=[\hat{\mathbf{r}}(a);\mathbf{e}(a)]$ for all $a\in\mathcal{A}$
  \State Fit the XGBoost ensemble on $\{(\phi(a),f(a))\}_{a\in\mathcal{L}_t}$
         to obtain $\mu(\cdot),\sigma(\cdot)$
  \State $\mathcal{B}_t \leftarrow$ the $b$ unevaluated architectures maximizing
         $\mu(a)+\beta\sigma(a)$
  \State Evaluate $f(a)$ for $a\in\mathcal{B}_t$; \;
         $\mathcal{L}_{t+1}\leftarrow\mathcal{L}_t\cup\mathcal{B}_t$
\EndFor
\State $\Pi_T \leftarrow$ re-selection on $\mathcal{L}_T$; refit the ensemble on
       $\mathcal{L}_T$ to obtain $\mu_{\mathrm{final}}(\cdot)$
\State \Return the $100$ architectures of $\mathcal{A}$ with the largest
       $\mu_{\mathrm{final}}(a)$
\end{algorithmic}
\end{algorithm}

\noindent
The loop consumes $n_0+Tb\le B$ true evaluations, so the budget is never exceeded;
the floor in $b$ leaves at most one evaluation unspent. The ranking returned is
produced by a final refit on the complete labeled set $\mathcal{L}_T$, after the
last batch has been evaluated.

\FloatBarrier
\section{Experiments}
\label{sec:experiments}

\subsection{Experimental Setup}
\label{sec:setup}

\paragraph{Benchmarks.}
Our main evaluation uses NAS-Bench-201~\citep{dong2020nasbench201}, a tabular
benchmark enumerating $15{,}625$ architectures generated by assigning one of five
operations to each of six cell edges, with recorded accuracies on CIFAR-10 and
CIFAR-100. Because every accuracy is a table lookup, no network is trained at any
point and results are exactly reproducible.
Section~\ref{sec:nb101} repeats the comparison on
NAS-Bench-101~\citep{ying2019nasbench101}, and the transferability study
(Section~\ref{sec:transferability}) additionally uses ImageNet16-120.
Proxy values are taken unmodified from
NAS-Bench-Suite-Zero~\citep{nasbenchsuitezero}.
A single \zaps{} run at $B=200$ completes in roughly five seconds on CPU.

\paragraph{Proxies.}
We use all $13$ proxies released by NAS-Bench-Suite-Zero, spanning deterministic
(\texttt{params}, \texttt{flops}, \texttt{l2\_norm}, \texttt{synflow}), activation
(\texttt{nwot}, \texttt{jacov}, \texttt{epe\_nas}, \texttt{zen}, \texttt{plain}) and
gradient (\texttt{fisher}, \texttt{snip}, \texttt{grad\_norm}, \texttt{grasp})
families. Individually their Spearman correlation with accuracy on
NAS-Bench-201 / CIFAR-10 ranges from $-0.27$ to $0.78$;
Table~\ref{tab:proxies} gives the breakdown, and
Figure~\ref{fig:proxy_corr} their pairwise redundancy.

\paragraph{Metrics.}
We report \patk{} (Eq.~\ref{eq:p100}), the percentage of the true top-100
architectures a method actually evaluated within its budget (higher is better), and
regret, the relative accuracy gap between the global optimum and the best
architecture it evaluated (lower is better).
As a complementary diagnostic specific to \zaps{}, we report the Spearman
correlation between its surrogate's predicted ranking and the ground-truth ranking
over all $15{,}625$ architectures; this quantity is defined only for methods that
rank the whole space and is therefore not used for comparison.
Unless stated otherwise, every number is a mean over $200$ independent random
seeds.

\paragraph{Baselines.}
We compare against two families. \emph{Search algorithms}: Random Search (RS),
Local Search (LS), regularized evolution (REA)~\citep{real2019regularized},
BANANAS~\citep{white2021bananas}, and TPE~\citep{bergstra2011tpe}, each granted the
same budget $B$ of true evaluations and run over the same $200$ seeds.
\emph{Zero-cost proxy combinations} (Section~\ref{sec:proxybaselines}): the best
single proxy, \texttt{synflow} used without calibration, the rank product, and a
Borda aggregation over the three proxies recommended by~\citet{rankproduct}.
These rank the space from proxy scores alone; to
spend the same budget as the search algorithms, each then trains the top $B$
architectures of its own ranking.

\subsection{Baseline Implementations and Conventions}
\label{sec:baselines}

All baselines are our own implementations. Table~\ref{tab:baselines} records, for
each, the settings we adopted and the behavioral check we used to verify it.
The global optima used to compute regret are $91.57\,\%$ (CIFAR-10),
$73.26\,\%$ (CIFAR-100) and $47.33\,\%$ (ImageNet16-120) on NAS-Bench-201, and
$94.72\,\%$ on NAS-Bench-101.
Two conventions apply throughout and are worth stating explicitly, since they
affect comparability.

\paragraph{Budget counts distinct architectures.}
Re-evaluating an architecture already seen costs nothing, since its accuracy is
known. This rule favours REA, whose mutations increasingly return architectures it
has already evaluated: charging every proposal would cost it $19.3\,\%$ of its
budget at $B=200$ and $41.6\,\%$ at $B=500$.

\paragraph{Projection onto the search space.}
NAS-Bench-101 encodes a cell in $56$ bits, i.e.\ $2^{56}$ possible vectors; its
validity constraints (connectivity, at most nine edges, a canonical form) leave
$423{,}624$ distinct architectures, and NAS-Bench-Suite-Zero provides proxy values
for a subset of $9{,}781$ of them, on which our entire evaluation is run. A mutated
encoding therefore almost never falls in this pool and must be projected back onto
it. The mutant lies at Hamming distance one from its parent, which is its nearest
neighbour in the pool in $97\,\%$ of cases, so the projection excludes it: REA then
draws the child uniformly among the eight nearest architectures, and TPE takes the
nearest one not yet evaluated. Without that exclusion a nearest-neighbour projection
returns the parent almost every time, so that $98\,\%$ of REA's mutations fall back
on a random draw and its \patk{} drops to that of Random Search. Even with it, about
a third of REA's mutations reach an architecture already evaluated and are replaced
by a random draw.

\begin{table}[H]
\centering
\footnotesize
\setlength{\tabcolsep}{4pt}
\caption{Baseline settings and the behavioral check used to verify each
implementation.}
\label{tab:baselines}
\adjustbox{max width=\linewidth}{%
\begin{tabular}{lp{0.36\linewidth}p{0.34\linewidth}}
\toprule
\textbf{Method} & \textbf{Settings} & \textbf{Verification} \\
\midrule
Random Search & sampling without replacement &
  matches $B\!\times\!100/N$ within $0.2$\,pp \\
Local Search & one-operation neighbourhood, steepest ascent, random restart on
  no improvement &
  $91.34\,\%$ at $B=200$, slightly below the $\approx\!91.5\,\%$ usually
  reported \\
REA & population $50$, tournament $10$, FIFO removal, one-operation mutation &
  $91.47\,\%$ at $B=200$, matching published values \\
BANANAS & path encoding, $5\times$ ten-layer MLP, independent Thompson sampling,
  mutation-based candidates &
  attains the global optimum by $B=500$ \\
TPE & $\gamma=0.25$, $64$ candidates, discrete KDE per position &
  \patk{} monotone in $B$ on both datasets \\
\bottomrule
\end{tabular}}
\end{table}

\subsection{Hyperparameters}
\label{sec:hp}

Table~\ref{tab:hp} lists every hyperparameter of \zaps{}. All values are held fixed
across datasets, budgets, and search spaces; none is tuned per experiment. Every
reported figure averages the seeds $0,\dots,199$, used identically by \zaps{} and
by every search baseline; the proxy-combination schemes of
Section~\ref{sec:proxybaselines} are deterministic and are run once.

\begin{table}[H]
\centering
\footnotesize
\setlength{\tabcolsep}{6pt}
\caption{Complete hyperparameter settings for \zaps{}.}
\label{tab:hp}
\adjustbox{max width=\linewidth}{%
\begin{tabular}{lll}
\toprule
\textbf{Stage} & \textbf{Hyperparameter} & \textbf{Value} \\
\midrule
\multirow{2}{*}{\pfit{}}
  & Redundancy threshold $\tau$          & $0.85$ \\
  & Target subset size $K$               & $6$ \\
\midrule
\multirow{4}{*}{Initialization}
  & Initialization budget $n_0$          & $\max(10,\lfloor B/4 \rfloor)$ \\
  & Exploitation fraction                & top $30\,\%$ by proxy rank \\
  & Clusters per phase                   & $n_0/2$ \\
  & K-means restarts / max iterations    & $3$ / $100$ \\
\midrule
\multirow{5}{*}{Re-selection}
  & Bootstrap resamples $B_{\mathrm{boot}}$ & $3$ \\
  & Minimum subset size $k_{\min}$       & $4$ \\
  & Maximum subset size $k_{\max}$       & $10$ \\
  & Target size $k$                      & $\max(k_{\min},\min(k_{\max},\lfloor|\mathcal{L}_t|/15\rfloor+4))$ \\
  & Redundancy threshold $\tau$          & $0.85$ \\
\midrule
\multirow{9}{*}{Surrogate}
  & Ensemble members $B_{\mathrm{ens}}$  & $3$ \\
  & Tree method / objective              & histogram / squared error \\
  & Trees per regressor                  & $50$ \\
  & Maximum tree depth                   & $5$ \\
  & Learning rate                        & $0.05$ \\
  & Row subsample                        & $0.8$ \\
  & Column subsample per tree            & $0.8$ \\
  & Minimum child weight                 & $3$ \\
  & $L_1$ regularization $\alpha$        & $0.1$ \\
\midrule
\multirow{3}{*}{Active loop}
  & Exploration weight $\beta$           & $0.5$ \\
  & Iterations $T$                       & $2$ \\
  & Batch size $b$                       & $\lfloor(B-n_0)/T\rfloor$ \\
\bottomrule
\end{tabular}}
\end{table}

\FloatBarrier
\subsection{Main Results}
\label{sec:main}

\begin{figure}[H]
\captionsetup{font=footnotesize,skip=2pt}
\centering
\includegraphics[width=\linewidth]{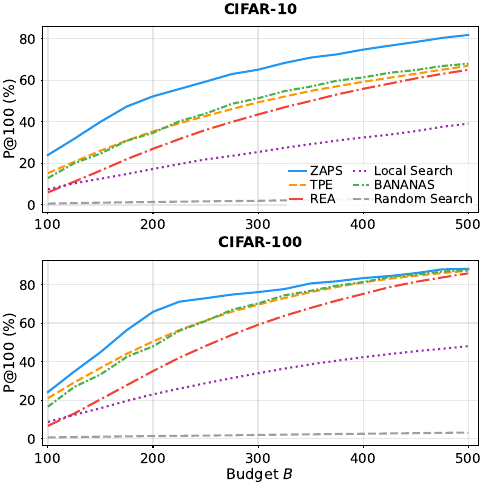}
\caption{\patk{} vs.\ budget $B$ on NAS-Bench-201 (top: CIFAR-10, bottom:
CIFAR-100); means over $200$ seeds.}
\label{fig:main}
\end{figure}

\begin{figure}[H]
\captionsetup{font=footnotesize,skip=2pt}
\centering
\includegraphics[width=\linewidth]{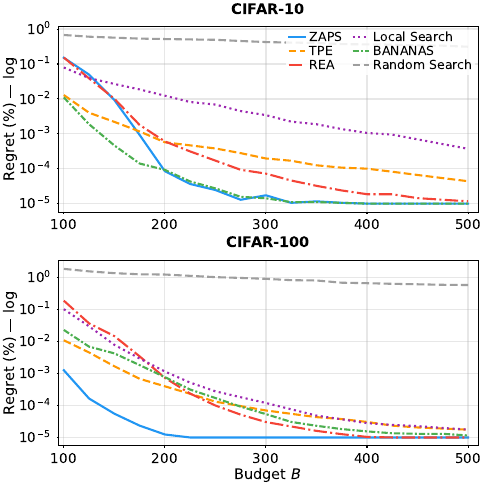}
\caption{Regret vs.\ budget $B$, same setting; geometric means over $200$
seeds, floored at $10^{-5}$.}
\label{fig:main_regret}
\end{figure}

Table~\ref{tab:budgets} reports performance at $B=200$ and $B=300$;
Figures~\ref{fig:main} and~\ref{fig:main_regret} trace the full
budget range.

\begin{table}[H]
\centering
\footnotesize
\setlength{\tabcolsep}{5pt}
\caption{Comparison on NAS-Bench-201 at $B=200$ and $B=300$ evaluations.
All methods are scored identically: \patk{} counts the true top-100 architectures
among those the method actually evaluated. Values are
mean${\scriptstyle\,\pm\,}$standard deviation over $200$ independent seeds.
Best value per column and budget in \textbf{bold}.
Every \patk{} gap between \zaps{} and every baseline is significant at $p<0.01$
(one-sided Mann--Whitney $U$); effect sizes and the one case where significance
does not imply practical relevance are discussed in the text.}
\label{tab:budgets}
\adjustbox{max width=\linewidth}{%
\begin{tabular}{cl cccc}
\toprule
& & \multicolumn{2}{c}{\textbf{CIFAR-10}}
& \multicolumn{2}{c}{\textbf{CIFAR-100}} \\
\cmidrule(lr){3-4}\cmidrule(lr){5-6}
$B$ & Method & \patk{} (\%) & Regret (\%) & \patk{} (\%) & Regret (\%) \\
\midrule
\multirow{6}{*}{$200$}
  & RS               & $1.4\pmstd{1.1}$   & $0.66\pmstd{0.26}$ & $1.4\pmstd{1.1}$   & $1.66\pmstd{0.88}$ \\
  & LS               & $17.3\pmstd{9.1}$  & $0.25\pmstd{0.23}$ & $23.0\pmstd{13.1}$ & $0.30\pmstd{0.51}$ \\
  & REA              & $27.0\pmstd{7.4}$  & $0.11\pmstd{0.15}$ & $35.1\pmstd{9.7}$  & $0.05\pmstd{0.10}$ \\
  & BANANAS          & $34.7\pmstd{10.6}$ & $0.06\pmstd{0.12}$ & $48.0\pmstd{11.7}$ & $0.03\pmstd{0.05}$ \\
  & TPE              & $35.4\pmstd{15.0}$ & $0.12\pmstd{0.18}$ & $50.5\pmstd{16.9}$ & $0.06\pmstd{0.19}$ \\
  & \zaps{}          & $\mathbf{52.3}\pmstd{6.1}$ & $\mathbf{0.05}\pmstd{0.09}$
                     & $\mathbf{65.8}\pmstd{11.0}$ & $\mathbf{0.02}\pmstd{0.14}$ \\
\midrule
\multirow{6}{*}{$300$}
  & RS               & $2.0\pmstd{1.4}$   & $0.59\pmstd{0.25}$ & $1.9\pmstd{1.4}$   & $1.36\pmstd{0.80}$ \\
  & LS               & $25.5\pmstd{10.1}$ & $0.17\pmstd{0.18}$ & $34.0\pmstd{13.1}$ & $0.12\pmstd{0.29}$ \\
  & REA              & $43.5\pmstd{8.8}$  & $0.05\pmstd{0.13}$ & $59.1\pmstd{7.6}$  & $\mathbf{0.00}\pmstd{0.01}$ \\
  & BANANAS          & $51.3\pmstd{7.5}$  & $\mathbf{0.01}\pmstd{0.04}$ & $70.2\pmstd{8.8}$ & $0.01\pmstd{0.04}$ \\
  & TPE              & $49.4\pmstd{18.3}$ & $0.09\pmstd{0.17}$ & $69.5\pmstd{16.0}$ & $0.03\pmstd{0.14}$ \\
  & \zaps{}          & $\mathbf{65.1}\pmstd{5.8}$ & $0.01\pmstd{0.05}$
                     & $\mathbf{76.1}\pmstd{4.5}$ & $\mathbf{0.00}\pmstd{0.00}$ \\
\bottomrule
\end{tabular}}
\end{table}

At $B=200$, \zaps{} recovers $52.3\,\%$ of the true top-100 on CIFAR-10, which is
$1.5\times$ the score of the strongest baseline, TPE ($35.4\,\%$), and
$16.9$\,pp above it. The margin is comparable on CIFAR-100, where \zaps{} reaches
$65.8\,\%$ against $50.5\,\%$ for TPE.

Beyond the mean, \zaps{} is also the most \emph{stable} method: at $B=200$ its
standard deviation is $6.1$\,pp on CIFAR-10 against $15.0$ for TPE and $10.6$ for
BANANAS, and $11.0$\,pp on CIFAR-100 against $16.9$ and $11.7$. Under a single-shot
budget, where a practitioner cannot average over seeds, this matters as much as the
mean.

\paragraph{Behaviour across budget regimes.}
The advantage of \zaps{} is largest where evaluations are scarce and narrows as they
accumulate (Figure~\ref{fig:main}). On CIFAR-10 it is stable, from $+16.9$\,pp over
the best baseline at $B=200$ to $+13.8$\,pp at $B=500$. On CIFAR-100 it decays
markedly, from $+15.4$\,pp at $B=200$ to a negligible $+0.6$\,pp at $B=500$, where
BANANAS, TPE and REA have nearly caught up with \zaps{}. This is what the design
predicts: the zero-cost prior is informative from the first iteration, but its
marginal value diminishes once enough true labels accumulate for a surrogate to be
fitted without it. Even at $B=500$ on CIFAR-100, where the gap has nearly closed, \zaps{}
remains the most stable method: its standard deviation is $3.7$\,pp,
against $11.1$ for TPE and $5.2$ for BANANAS.

\paragraph{Statistical significance.}
With $200$ seeds per cell we test these gaps rather than assert them, using a
one-sided Mann--Whitney $U$ test on per-seed \patk{} values and Cliff's $\delta$ as
effect size ($|\delta|<0.15$ negligible, above $0.47$ large). At $B=200$ the
advantage is unambiguous on both datasets ($p<10^{-24}$ against every baseline),
with a large effect against the strongest of them ($\delta=0.69$ on CIFAR-10,
$0.60$ on CIFAR-100, both versus TPE), and remains large at $B=300$ on CIFAR-10
($p<10^{-22}$, and $\delta=0.57$ versus TPE, the weakest effect). On CIFAR-100 at
$B=300$, however, the gap to TPE is still detectable ($p<0.01$) while $\delta$ has
fallen to a negligible $0.14$; against BANANAS, the best on average, it remains
medium ($\delta=0.45$):
detectability and practical relevance part company, so we report both rather than
$p$ alone. On regret the tests are less favourable---\zaps{} does not separate from
BANANAS on CIFAR-10 at $B=200$ ($p=0.25$), a point Section~\ref{sec:conclusion}
takes up.

\paragraph{Baseline implementations.}
Every baseline is our own re-implementation, since no reference code targets these
benchmarks in a directly usable form. Because a comparison is only as meaningful as
its weakest baseline, we audited each against its source publication and against a
behavioral check---BANANAS reproduces the global CIFAR-10 optimum by $B=500$ as in
the original work, and Random Search matches its hypergeometric expectation to
within $0.2$\,pp---and we report those checks rather than asking the reader to take
faithfulness on trust. Two conventions affect comparability and are stated
explicitly in Section~\ref{sec:baselines}: the budget counts \emph{distinct}
architectures, and the projection of a mutated encoding back onto NAS-Bench-101
must exclude the parent, without which a nearest-neighbour projection would
reduce REA to random sampling.

\begin{figure}[H]
\captionsetup{font=scriptsize,skip=2pt}
\centering
\begin{minipage}[t]{\linewidth}
\centering
\includegraphics[width=\linewidth]{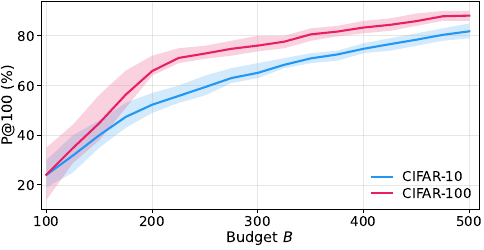}
\caption{Convergence of \zaps{} alone: \patk{} vs.\ budget $B$ on CIFAR-10 and
CIFAR-100; means and interquartile bands over $200$ seeds.}
\label{fig:conv}
\end{minipage}
\end{figure}

\begin{figure}[H]
\captionsetup{font=scriptsize,skip=2pt}
\centering
\begin{minipage}[t]{\linewidth}
\centering
\includegraphics[width=\linewidth]{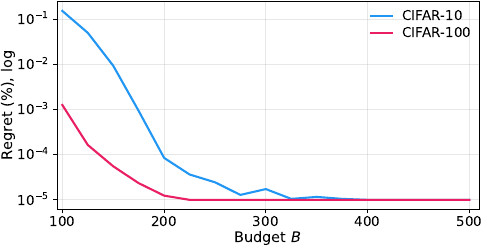}
\caption{Regret of \zaps{} alone, same setting; geometric means over $200$
seeds, floored at $10^{-5}$.}
\label{fig:conv_regret}
\end{minipage}
\end{figure}

\begin{figure}[H]
\captionsetup{font=scriptsize,skip=2pt}
\centering
\begin{minipage}[t]{\linewidth}
\centering
\includegraphics[width=\linewidth]{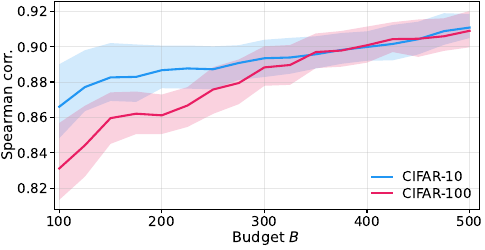}
\caption{Spearman correlation between the surrogate ranking and the true
ranking of all $15{,}625$ architectures of NAS-Bench-201, on CIFAR-10 and
CIFAR-100; means and interquartile bands over $200$ seeds.}
\label{fig:spearman}
\end{minipage}
\end{figure}

\paragraph{Ranking quality.}
The surrogate is accurate globally, not only near the optimum: from $B=100$ onward
it attains a Spearman correlation of $\rho\ge 0.86$ on CIFAR-10 and $\rho\ge 0.83$
on CIFAR-100 against the full ground-truth ranking, rising to $\rho=0.89$ at
$B=200$ (Figure~\ref{fig:spearman}).
Figures~\ref{fig:conv} and~\ref{fig:conv_regret} trace \zaps{}'s own
convergence across the budget range, on CIFAR-10 and CIFAR-100.

\subsection{Comparison with Zero-Cost Proxy Combination}
\label{sec:proxybaselines}

Since our premise is that proxies must be \emph{selected} rather than merely
aggregated, we compare against the proxy-combination schemes \pfit{} is designed to
improve upon. A zero-cost method produces its ranking without training anything, so
that ranking cannot be set directly against a method that has spent $B$ evaluations.
We therefore give each scheme the budget it would consume in practice: it ranks the
space from proxy scores, then trains the top $B$ of that ranking. Every row of
Table~\ref{tab:proxybaselines} thus spends
exactly $B=200$ true evaluations, and
\patk{} counts for all of them how much of the true top-100 was actually trained.
The proxy schemes are deterministic, so no standard deviation is reported.

\begin{table}[H]
\centering
\small
\setlength{\tabcolsep}{6pt}
\caption{Zero-cost proxy schemes on NAS-Bench-201, each spending $B=200$ true
evaluations: the scheme ranks the space from proxy scores, then trains its top
$200$; \patk{} counts the true top-100 among the architectures trained.
\emph{Best single} is chosen by highest Spearman $\rho$ against accuracy, using
oracle knowledge unavailable to a practitioner. The last row aggregates the three
proxies recommended by~\citet{rankproduct} by summing their normalized ranks
(Borda count); we use this rather than their pairwise majority vote, which does not
in general induce a total order over the space. For reference, \zaps{} reaches
$52.3$ and $65.8\,\%$ at the same budget.}
\label{tab:proxybaselines}
\adjustbox{max width=\linewidth}{%
\begin{tabular}{l cc cc}
\toprule
& \multicolumn{2}{c}{\textbf{CIFAR-10}} & \multicolumn{2}{c}{\textbf{CIFAR-100}} \\
\cmidrule(lr){2-3}\cmidrule(lr){4-5}
Method ($B=200$) & \patk{} (\%) & Regret (\%) & \patk{} (\%) & Regret (\%) \\
\midrule
\texttt{nwot} (best single, oracle)   & $0.0$  & $0.91$ & $7.0$  & $1.50$ \\
\texttt{synflow} (fixed)              & $22.0$ & $\mathbf{0.00}$ & $36.0$ & $0.03$ \\
Rank product, all 13                  & $0.0$  & $1.34$ & $6.0$  & $1.50$ \\
Rank product, positive only           & $0.0$  & $0.91$ & $8.0$  & $1.50$ \\
Borda, \{\texttt{synflow},\texttt{jacov},\texttt{snip}\} & $0.0$ & $1.06$ & $5.0$ & $1.17$ \\
\bottomrule
\end{tabular}}
\end{table}

The strongest scheme, \texttt{synflow} used alone, recovers $22.0\,\%$ of the true
top-100 on CIFAR-10 and $36.0\,\%$ on CIFAR-100 for the same $200$ evaluations.
That places it above Random Search and Local Search on both datasets, and above REA
on CIFAR-100, but below the surrogate-based baselines BANANAS and TPE
(Table~\ref{tab:budgets}): a well-chosen proxy is a strong starting point, not a
substitute for search. Two further observations bear on the motivation
for \pfit{}, and neither involves \zaps{}.

\emph{Naive combination is worse than no combination.} Every aggregation scheme
falls far short of \texttt{synflow} used alone: on CIFAR-10 the rank product over
all thirteen proxies and the Borda aggregation over the three proxies
of~\citet{rankproduct} both reach $0.0\,\%$ against $22.0\,\%$ for
\texttt{synflow}, and restricting the product to positively correlated proxies does
not rescue it. Selecting the single proxy by Spearman $\rho$ fares no better---it
picks \texttt{nwot} ($\rho=0.775$, the highest of the thirteen), which also reaches
$0.0\,\%$: a proxy can order the bulk of the space well while being uninformative
exactly where a NAS method operates, which is why we report \patk{} rather than the
Spearman correlation common in the zero-cost literature. Aggregating without
accounting for redundancy destroys the signal each proxy carried on its own, because
correlated proxies reinforce one another's errors instead of averaging them out---the
failure mode \pfit{} is built to avoid, and which Section~\ref{sec:ablation} measures
again inside the full pipeline.

\subsection{Ablation Study}
\label{sec:ablation}

\begin{figure}[H]
\centering
\includegraphics[width=\linewidth]{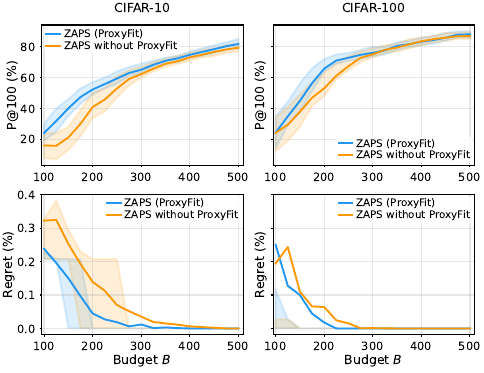}
\caption{Ablation of \pfit{} across the full budget range on CIFAR-10 and
CIFAR-100: the \pfit{} subset of six proxies against the full pool of
thirteen. Lines are means over $200$ seeds and shaded bands interquartile
ranges. Table~\ref{tab:ablation} reports the values at $B=200$.}
\label{fig:ablations}
\end{figure}

\begin{figure}[H]
\centering
\includegraphics[width=\linewidth]{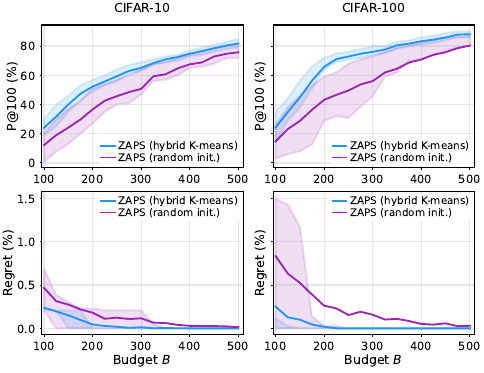}
\caption{Ablation of the initialization across the full budget range on
CIFAR-10 and CIFAR-100: hybrid K-means initialization against uniform random
sampling. Lines are means over $200$ seeds and shaded bands interquartile
ranges. Table~\ref{tab:ablation} reports the values at $B=200$.}
\label{fig:ablations_init}
\end{figure}

We isolate the contribution of the two components that shape the surrogate's input:
the anti-redundant proxy selection of \pfit{} and the structured initialization.
Table~\ref{tab:ablation} reports both at $B=200$;
Figures~\ref{fig:ablations} and~\ref{fig:ablations_init} show the
effect across the budget range.

\paragraph{Structured initialization is the dominant factor.}
Replacing hybrid K-means by uniform random sampling costs $15.6$\,pp of \patk{} on
CIFAR-10 and $22.6$\,pp on CIFAR-100. More strikingly, it inflates the standard
deviation from $6.1$ to $17.3$\,pp on CIFAR-10 and quadruples regret
($0.045\to0.181$). Without a structured seed the outcome of a run is largely
decided by whether the initial sample happened to land near a good region---exactly
the fragility a small budget cannot absorb. Note that this variant still uses the
\pfit{} subset, so the figure isolates the seeding strategy alone.

\paragraph{Anti-redundancy filtering contributes consistently.}
Seeding the hybrid K-means with all thirteen proxies instead of the \pfit{} subset
costs $11.4$\,pp on CIFAR-10 and $12.8$\,pp on CIFAR-100, and raises the standard
deviation from $6.1$ to $10.8$\,pp. Only the initialization differs here: in both
cases the surrogate is fed the subset returned by the online re-selection, so the
comparison isolates \pfit{} itself. Redundant proxies distort the aggregate score
that defines the exploitation region, so the initial labels concentrate where a
signal counted several times over happens to point. Regret follows, tripling
from $0.045\,\%$ to $0.139\,\%$ on CIFAR-10. This is the same failure mode observed
in Section~\ref{sec:proxybaselines}, where naive rank aggregation underperformed
the single best proxy---here measured inside the full pipeline rather than on the
raw proxy rankings.

\begin{table}[H]
\centering
\footnotesize
\setlength{\tabcolsep}{6pt}
\caption{Component ablation on NAS-Bench-201 at $B=200$, over $200$ seeds.
Each row removes a single component from the full pipeline. Both variants alter
only the initialization---the first seeds the hybrid K-means with all thirteen
proxies, the second replaces it by uniform sampling---while the online
re-selection and the surrogate are unchanged.}
\label{tab:ablation}
\adjustbox{max width=\linewidth}{%
\begin{tabular}{l cccc}
\toprule
& \multicolumn{2}{c}{\textbf{CIFAR-10}} & \multicolumn{2}{c}{\textbf{CIFAR-100}} \\
\cmidrule(lr){2-3}\cmidrule(lr){4-5}
Variant & \patk{} (\%) & Regret (\%) & \patk{} (\%) & Regret (\%) \\
\midrule
\zaps{} (full pipeline)            & $\mathbf{52.3}\pmstd{6.1}$  & $\mathbf{0.045}$ & $\mathbf{65.8}\pmstd{11.0}$ & $\mathbf{0.018}$ \\
\quad w/o \pfit{} (init.\ on all 13) & $40.8\pmstd{10.8}$ & $0.139$ & $53.0\pmstd{15.4}$ & $0.064$ \\
\quad w/o K-means (random init.)   & $36.7\pmstd{17.3}$ & $0.181$ & $43.2\pmstd{21.4}$ & $0.261$ \\
\bottomrule
\end{tabular}}
\end{table}

\begin{figure}[H]
\centering
\includegraphics[width=\linewidth]{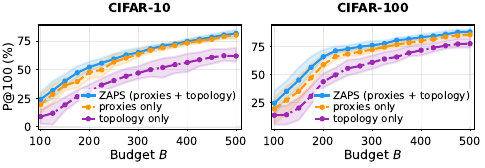}
\\[6pt]
\includegraphics[width=\linewidth]{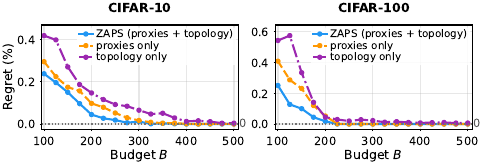}
\captionsetup{font=scriptsize,skip=2pt}
\caption{Fusion ablation, $200$ seeds; only the surrogate's input differs.
Topology-only never closes its gap; proxies-only converges to the full
pipeline. Regret on a linear scale, since \zaps{} reaches exactly zero.}
\label{fig:fusion}
\end{figure}

\paragraph{Do the two feature blocks actually complement each other?}
Our central hypothesis is that proxy ranks and topological encodings carry
different information. We test it directly by cutting the surrogate's input in two,
leaving the rest of the pipeline untouched so that any difference is attributable
to the features alone: \emph{proxies only} keeps the re-selected proxy ranks,
\emph{topology
only} keeps the $30$-dimensional one-hot encoding (Table~\ref{tab:fusion};
curves in Figure~\ref{fig:fusion}).

Both blocks contribute significantly, but far from symmetrically. Removing the
topology costs $4.5$\,pp on CIFAR-10 and $7.1$\,pp on CIFAR-100 ($p<10^{-7}$);
removing the proxies costs $21.2$ and $22.4$\,pp ($p<10^{-53}$). The proxies are
the dominant signal, which is consistent with the rest of the pipeline, since
\pfit{} and the initialization both operate on proxy ranks.

Two observations qualify this. First, the topology-only variant reaches $31.1\,\%$
on CIFAR-10---\emph{below} BANANAS ($34.7\,\%$), itself a surrogate over
architecture encodings. Stripped of its proxies, our pipeline is an ordinary
encoding-based surrogate; the proxies are what lift it to $52.3\,\%$. Second, the
contribution of the topology shrinks with the budget, from $+4.6$\,pp at $B=100$ to
$+1.5$\,pp at $B=500$ on CIFAR-10, whereas the gap to the topology-only variant
never falls below $15$\,pp at any budget. A surrogate deprived of proxies never
catches up;
one deprived of topology largely does, once enough labels accumulate for it to
learn the structure directly.

\begin{table}[H]
\centering
\footnotesize
\setlength{\tabcolsep}{6pt}
\caption{Fusion ablation on NAS-Bench-201 at $B=200$, $200$ seeds. Only the
surrogate's input changes; \pfit{}, initialization, re-selection and UCB
acquisition are identical across rows. The proxy block is the online
re-selection's output, whose size grows with the labeled set: $7$ proxies at the
first fit, $10$ thereafter.}
\label{tab:fusion}
\begin{tabular}{l c cc}
\toprule
Surrogate input & Dim. & \textbf{CIFAR-10} & \textbf{CIFAR-100} \\
\midrule
Proxy ranks $+$ topology & $37$--$40$ & $\mathbf{52.3}\pmstd{6.1}$ & $\mathbf{65.8}\pmstd{11.0}$ \\
\quad proxy ranks only   & $7$--$10$ & $47.8\pmstd{8.4}$ & $58.7\pmstd{14.0}$ \\
\quad topology only      & $30$ & $31.1\pmstd{8.0}$ & $43.4\pmstd{11.0}$ \\
\bottomrule
\end{tabular}
\end{table}

\subsection{A Second Search Space: NAS-Bench-101}
\label{sec:nb101}

NAS-Bench-201 is close to saturated, so we repeat the comparison on the harder
NAS-Bench-101~\citep{ying2019nasbench101}: cells vary in node count, the space
holds $423{,}624$ architectures (we use the $9{,}781$ annotated by
NAS-Bench-Suite-Zero), and the encoding has $56$ dimensions rather than $30$.
Local Search and BANANAS are omitted, as their neighbourhood and mutation
operators do not transfer directly to variable-topology graphs.

\begin{figure}[H]
\centering
\includegraphics[width=\linewidth]{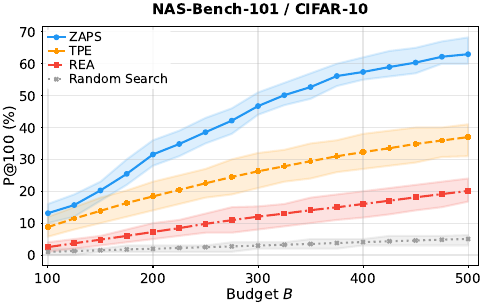}
\\[4pt]
\includegraphics[width=\linewidth]{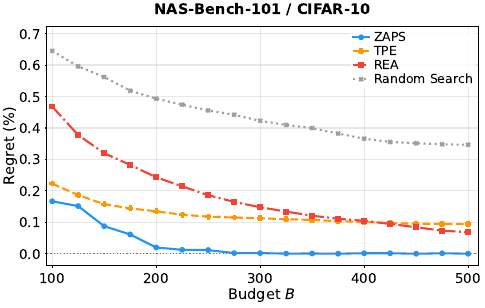}
\captionsetup{font=scriptsize,skip=2pt}
\caption{\patk{} (top) and regret (bottom) on NAS-Bench-101 / CIFAR-10; means over
$200$ seeds, with interquartile bands for \patk{}.}
\label{fig:nb101}
\end{figure}

\zaps{} leads at every budget (Table~\ref{tab:nb101}, Figure~\ref{fig:nb101}):
$31.5\,\%$ \patk{} at $B=200$ against $18.4\,\%$ for TPE, and $63.0\,\%$ at
$B=500$ against $37.0\,\%$. Unlike on CIFAR-100 (Section~\ref{sec:main}), the gap
\emph{widens} with the budget, from $+4.3$\,pp at $B=100$ to $+26.0$\,pp at
$B=500$: where no method can exhaust the space, the proxy prior keeps paying off.
The surrogate's global Spearman correlation is lower ($0.62$ at $B=200$, vs $0.89$
on NAS-Bench-201): \zaps{} models this space less well overall, yet still finds
its top region.

\begin{table}[H]
\centering
\footnotesize
\setlength{\tabcolsep}{3pt}
\caption{\patk{} (\%) on NAS-Bench-101 / CIFAR-10, $200$ seeds, unified metric.}
\label{tab:nb101}
\begin{tabular}{l cccc}
\toprule
Method & $B=100$ & $B=200$ & $B=300$ & $B=500$ \\
\midrule
Random Search & $1.0$ & $2.0$  & $3.0$  & $5.0$ \\
REA           & $2.5$ & $7.2$  & $12.0$ & $20.1$ \\
TPE           & $8.8$ & $18.4$ & $26.3$ & $37.0$ \\
\midrule
\zaps{}       & $\mathbf{13.1}$ & $\mathbf{31.5}$ & $\mathbf{46.7}$ & $\mathbf{63.0}$ \\
\bottomrule
\end{tabular}
\end{table}

\subsection{Transferability of the Selected Proxies}
\label{sec:transferability}

\begin{figure}[H]
\centering
\includegraphics[width=\linewidth]{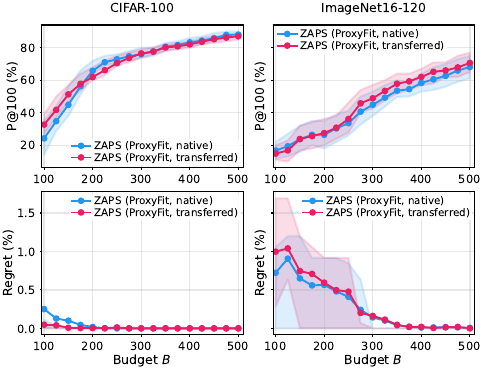}
\caption{Cross-dataset transfer: the subset calibrated on CIFAR-10, applied
unchanged to CIFAR-100 and ImageNet16-120 (\emph{transferred}), against one
calibrated on the target itself (\emph{native}). Values in
Table~\ref{tab:transfer}.}
\label{fig:transfer_dataset}
\end{figure}

\begin{figure}[H]
\centering
\includegraphics[width=\linewidth]{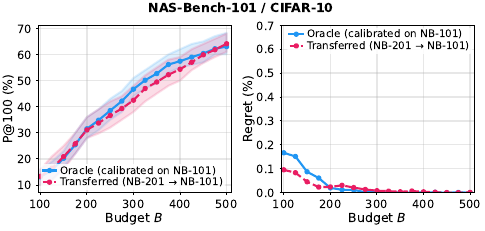}
\caption{Cross-space transfer: the subset calibrated on NAS-Bench-201, applied
unchanged to NAS-Bench-101 (\emph{transferred}), against one calibrated on
NAS-Bench-101 (\emph{oracle}). They differ by at most $4.2$\,pp in \patk{} and
$0.07$\,pp in regret.}
\label{fig:transfer_space}
\end{figure}

\pfit{} requires ground-truth accuracies, so a natural objection is that it must
be re-run---at cost---for every new dataset or space. It need not be: applying the
CIFAR-10 subset directly to CIFAR-100 and ImageNet16-120 costs essentially nothing.
Beyond the smallest budgets, the transferred subset stays close to one calibrated on the target dataset
(Figure~\ref{fig:transfer_dataset}),
with neither ahead systematically---on ImageNet16-120 it is even the better of the
two at most budgets. At the smallest budget, transfer is clearly ahead on
CIFAR-100, the target calibration apparently overfitting its offline estimate.
Transferring the
NAS-Bench-201 subset to the topologically different NAS-Bench-101 likewise tracks
an oracle calibrated directly on it (Table~\ref{tab:transfer}, Figure~\ref{fig:transfer_space}).

Proxy redundancy thus appears to be a property of the architectures themselves, not
the dataset, so \pfit{} can be treated as a reusable one-off calibration, removing
its cost from any subsequent search's budget and supporting the claim in
Section~\ref{sec:proxyfit} that ground-truth access does not drive our results.

\begin{table}[H]
\centering
\footnotesize
\setlength{\tabcolsep}{4pt}
\caption{\patk{} (\%) when \pfit{} is calibrated on the target dataset
(\emph{native}) versus transferred unchanged from CIFAR-10 (\emph{transferred}).
Mean over $200$ seeds. A transferred subset uses no ground-truth accuracy from the
target dataset.}
\label{tab:transfer}
\begin{tabular}{l cc c cc}
\toprule
& \multicolumn{2}{c}{\textbf{CIFAR-100}} & &
  \multicolumn{2}{c}{\textbf{ImageNet16-120}} \\
\cmidrule(lr){2-3}\cmidrule(lr){5-6}
$B$ & native & transferred & & native & transferred \\
\midrule
$100$ & $24.2$ & $\mathbf{32.7}$ & & $\mathbf{16.7}$ & $14.8$ \\
$200$ & $\mathbf{65.8}$ & $62.0$ & & $26.4$ & $\mathbf{27.5}$ \\
$300$ & $76.1$ & $\mathbf{76.5}$ & & $44.9$ & $\mathbf{48.9}$ \\
$500$ & $\mathbf{88.1}$ & $87.1$ & & $68.1$ & $\mathbf{70.7}$ \\
\bottomrule
\end{tabular}
\end{table}

\FloatBarrier
\section{Conclusion}
\label{sec:conclusion}

\zaps{} unifies zero-cost proxy selection and surrogate-based active learning
under a strict evaluation budget. On NAS-Bench-201 at $B=200$ it recovers
$52.3\,\%$ of the true top-100 on CIFAR-10 and $65.8\,\%$ on CIFAR-100, ahead of
every baseline by at least $15$\,pp and, on CIFAR-10, with less than half the
strongest's run-to-run standard deviation. The guiding hypothesis---proxy ranks and topological
encodings are complementary---holds, though asymmetrically: removing the proxies
costs three to five times what removing the topology does. The contribution is thus as much \emph{selecting} a non-redundant proxy subset as
fusing it with structure, and that selection transfers across datasets and spaces.

\paragraph{Limitations.}
Our contribution is narrower than ``better NAS''. \zaps{} excels at recovering the
top-100 \emph{set} under a constrained budget, but does not separate from BANANAS
on returning the single best CIFAR-10 architecture, where both reach the optimum
in about $78\,\%$ of seeds at $B=200$ and all by $B=400$ ($p=0.25$ on regret). The
advantage also has a budget window: on CIFAR-100 the \patk{} gap falls from
$+15.4$ to $+0.6$\,pp between $B=200$ and $B=500$---though not on NAS-Bench-101
(Section~\ref{sec:nb101}), verified on two spaces only. Finally, evaluation is
confined to tabular benchmarks with savings unverified for real training, and the
encoding assumes fixed cell topology.

\paragraph{Reproducibility statement.}
All experiments are table lookups on public benchmarks with no network training,
so the reported numbers are exactly reproducible from the accompanying artifacts.
Complete hyperparameters are listed in Section~\ref{sec:hp}; code, run scripts,
and per-seed results for all $200$ seeds and all budgets will be released
publicly, and are available from the authors on request in the meantime.

\paragraph{Ethics statement.}
This work lowers the computational cost of architecture selection and may thereby
broaden access to neural architecture design. We identify no direct negative
societal impact: the study uses only public benchmark data, involves no human
subjects or personally identifying information, and requires no GPU training.

\paragraph{AI use statement.}
A large language model (Claude, Anthropic) was used as an assistant throughout this
work, and we describe its role by task rather than in general terms.

\emph{Research idea and method design.} The four-stage pipeline, the
anti-redundancy criterion of \pfit{}, the two-phase initialization and the choice
to concatenate proxy ranks with topological encodings are the authors' own. The
model did not propose the method.

\emph{Implementation.} The authors wrote the original implementation of \zaps{} and
of the baselines. The model subsequently audited that code and identified the
following defects, all in the baselines: BANANAS used a one-hot rather than a path encoding
and optimized its acquisition over the full space rather than over mutations; REA
counted duplicate evaluations against its budget on NAS-Bench-201 and, on
NAS-Bench-101, used a projection that returned the parent in $99\,\%$ of mutations;
TPE contained a fatal name error on NAS-Bench-201 and lost its exploitation
component on NAS-Bench-101; one metric routine could loop indefinitely; and the
Random Search generator had been lost. The model wrote the corrections and the
replacement Random Search script, which the authors reviewed. All reported baseline
numbers come from the corrected implementations.

\emph{Experiment design.} The fusion ablation of Section~\ref{sec:ablation}---the
proxies-only and topology-only variants that test this paper's central
hypothesis---was proposed by the model after it observed that no existing experiment
isolated the contribution of the two feature blocks. The authors approved and ran it.
The model also proposed the significance testing and effect-size reporting.

\emph{Analysis and writing.} The model produced the figure-generation scripts, ran
the statistical tests, and drafted substantial portions of the text, which the
authors edited. Because a language model can restate a number incorrectly, every
numerical claim in this paper was subsequently re-derived from the result files by
a verification script rather than accepted as written. That audit was run
repeatedly as the results were finalized, and every discrepancy it surfaced---in
reported means, effect sizes, significance thresholds and figure captions---was
corrected against the result files. The verification scripts are released with the
code.

\emph{What the model did not do.} It did not generate data, run experiments without
authorization, or produce any result reported here that is not backed by a committed
CSV file. The authors take full responsibility for the contents of this paper.

\balance
\bibliographystyle{iclr2027_conference}
\bibliography{references}

\end{document}